\documentclass[11pt]{llncs}
\usepackage[english]{babel}
\usepackage[utf8]{inputenc}
\usepackage{epsfig,amsfonts,latexsym,amsmath,amssymb,color,axodraw4j,pstricks,graphicx,palatino,url,parskip,algorithm,algorithmic,tikz,array}

\begin{document}

\title{New Algorithmic Approaches to Point Constellation Recognition}

\author{Thomas Bourgeat\inst{1}, Julien Bringer\inst{2}, Herv\'e Chabanne\inst{2}, Robin Champenois\inst{1}, Jérémie Clément\inst{1,3}, Houda Ferradi\inst{1,2}, Marc Heinrich\inst{1}, Paul Melotti\inst{1}, David Naccache\inst{1}, Antoine Voizard\inst{1}}

\institute{%
\'Ecole normale sup\'erieure\\
\'Equipe de cryptographie, 45 rue d'Ulm, {\sc f}-75230 Paris {\sc cedex 05}, France\\
\email{given\_name.name@ens.fr}
\and
Morpho\\
11 boulevard Gallieni, {\sc f}-92130 Issy-les-Moulineaux, France.\\
\email{given\_name.name@morpho.com}
\and
Crocus Technology\\
4 place Robert Schuman, {\sc f}-38025 Grenoble Cedex, France \\
\email{jclement@crocus-technology.com}
}

\maketitle
\begin{abstract}
Point constellation recognition is a common problem with many pattern matching applications. Whilst useful in many contexts, this work is mainly motivated by fingerprint matching. Fingerprints are traditionally modelled as constellations of oriented points called {\sl minutiae}. The fingerprint verifier's task consists in comparing two point constellations. The compared constellations may differ by rotation and translation or by much more involved transforms such as distortion or occlusion.\smallskip

This paper presents three new constellation matching algorithms. The first two methods generalize an algorithm by Bringer and Despiegel. Our third proposal creates a very interesting analogy between mechanical system simulation and the constellation recognition problem.
%\textbf{Keywords :} Fingerprints, Vicinity, Minutiae, Spring Authentication
\end{abstract}

\section{Introduction}
Fingerprints are traditionally modelled as feature-sets called {\sl minutiae}. Each minutia $m_i$ is composed of two cartesian coordinates and an orientation $\{x_i, y_i, \theta_i\}$. Matching algorithms return a score expressing the similarity between the candidate minutiae set $M=\{m_1,\ldots,m_{\ell}\}$ and the template minutiae set $M'=\{m'_1,\ldots,m'_{\ell'}\}$. The main difficulty met while trying to compute such similarity scores is that many experimental parameters may affect the minutiae comparison task. The easiest obstacles to deal with are simple transforms such as translation and rotation but minutiae may also disappear (occlusion), appear {\sl ex nihilo} (noise) or be subject to nonlinear distortion. Most matching algorithms usually try to translate and rotate $M'$ to obtain an optimal superimposition with $M$ (while dealing with nonlinear transforms using a variety of heuristics). In general, matching algorithms use a $\{x_i, y_i, \theta_i\}$ representation standardized in ISO/IEC 19794-2 \cite{ISO19794}.\\

In \cite{BFVFRFMV} Bringer and Despiegel (BD) define the notion of {\sl vicinity}. Instead of using minutiae to compare $M$ and $M'$, \cite{BFVFRFMV} groups  minutiae into vicinities $V_i$. A vicinity $V_i$ is defined as a center minutia $m_i=\{x_i, y_i, \theta_i\}$ and all the minutiae $m_j$ whose distance to $m_i$ does not exceed a certain range. The advantage of this representation is that each vicinity $V_i$ carries its own coordinate system, whose center is $\{x_i, y_i\}$ and whose $x$-axis is $\theta_i$. Thereby, rotation and translation issues are naturally avoided. \\

This paper presents three new point constellation matching algorithms. The first two techniques generalize BD's algorithm. Section 2 recalls BD's ideas and notations, which will serve as a basis for our two first algorithms. Section 3 describes a technique based on {\sl second-order vicinities}, which are vicinities of vicinities. Second-order vicinities allow to extract more information from constellations by exploiting the information present in the relative distances separating vicinities. Section 4 describes {\sl missing minutiae analysis}, a technique allowing to extract useful information from... missing minutiae. Section 5, introduces a matching method that stems from a very interesting analogy between mechanical system simulation and the constellation recognition problem.

\section{Fingerprints, Minutiae and Vicinities}

\subsection{From Minutiae to Vicinities}

As we have already mentioned, a vicinity $V_i$ consists of all the minutiae $m_j$ whose distance from $m_i$ does not exceed some range $\rho$, {\sl i.e.}:

$$m_j \in V_i \Leftrightarrow \sqrt{(x_i-x_j)^2+(y_i-y_j)^2}<\rho$$

Each $V_i$ has its own coordinate system. $\{x_i,y_i\}$ is the center of this coordinate system and $\theta_i$ provides the orientation of $V_i$'s $x$-axis. All the $m_j\in V_i$ have their coordinates recomputed with respect to $\{x_i,y_i\}$. Now a fingerprint $M$ can be regarded as a set of vicinities instead of a set of minutiae. If $M$ contains $n$ minutiae then $M$ will also yield $n$ vicinities. To compare two vicinities (hereafter $A$ and $B$, respectively containing the minutiae $a_i$ and $b_j$) we will compare the $a_i$s to the $b_j$s pairwise. For each $a_i$ {\sl vs.} $b_j$ comparison a matching score will be computed. Here we use the simplified scoring formula:

\[s(a_i, b_j) = (x_{a_i} - x_{b_j})^2 + (y_{a_i} - y_{b_j})^2 + \frac{\sigma_x}{\sigma_\theta}(\theta_{a_i} - \theta_{b_j})^2\]

where $\sigma_x$ represents the variance of the position (we assume that $\sigma_x=\sigma_y$), and $\sigma_\theta$ is the variance of the orientation. Experimentally measured vicinities are used to tune these parameters.\\

Each pair of minutiae yields a score. A matrix containing all these scores is built. Then, the Hungarian algorithm \cite{THMAP} is applied to this matrix to find the best association between the minutiae of $A$ and $B$. The final matching score between the two vicinities will be computed as follows :

\[S_{A,B} = \sum_{f(a_i)=\{b_j\}} s(a_i, b_j) - (\mbox{NAR}(A,B) + \mbox{NAS}(A,B))K_{\mbox{{\scriptsize NA}}}\]

$$\mbox{where~}f(a_i) =
\left\{
	\begin{array}{ll}
		\{b_j\}  & \mbox{if~} a_i \mbox{~is associated to~} b_j \\
		0 & \mbox{otherwise}
	\end{array}
\right.$$

$\mbox{NAR}$ represents the number of minutiae of $A$ that do not have any association in $B$ and $\mbox{NAS}$ is the number of minutiae of $B$ that do not have any association in $A$. $K_{\mbox{{\scriptsize NA}}}$ is a penalty coefficient for non associated minutiae.\\

Once we have the scores corresponding to the comparisons between the template's and candidate's vicinities, we can compute the binary {\sl feature vector} which will represent our fingerprint.

\subsection{From Vicinities to the Binary Feature Vector}

To create a {\sl feature vector} that will represent a fingerprint we will need $N$ representative vicinities. This set of representative vicinities, denoted $\mbox{DB}_R$, contains $N$ vicinities $R_i$ that fulfill several conditions:\\

\begin{itemize}
\item The $R_i$s are not related to the main fingerprint database. The $R_i$s come from generated fingerprints or external databases.
\item The $R_i$s contain more than $\ell_{\mbox{{\scriptsize min}}}$ and less than $\ell_{\mbox{{\scriptsize max}}}$ minutiae. $\ell_{\mbox{{\scriptsize min}}}$ and $\ell_{\mbox{{\scriptsize max}}}$ are chosen so the $R_i$s are not too discriminative.
\item Each $R_i$ isn't similar to other $R_j$s (according to a certain threshold criterion).
\end{itemize}

The vector $V$, of length $N$, representing the candidate fingerprint is computed as follows:

\begin{itemize}
\item Extract all the vicinities (denoted $F_j$) from the candidate fingerprint.
\item For $1\leq i \leq N$ compute the matching scores $S(F_j,R_i)$
\item Given a certain threshold $t$, create the vector $V$:

\[V_i=\left\{
  \begin{array}{cl}
     1 & \mbox{~~if~}\exists j \in \{1,\ldots,n\} \mbox{ such as }S(F_j, R_i)<t \\
     0 & \mbox{~~otherwise}\\
  \end{array}
\right.\]
\end{itemize}

To compare two fingerprints, compute the hamming distance of their feature vectors.\\

This can be used both for authentication (compare a candidate fingerprint to a database of fingerprints, this database contains only the binary vectors of each fingerprint) and for identification (compare the candidate fingerprint to one template fingerprint to check the identity). This method has three main advantages :
\begin{itemize}
\item Avoiding rotation and translation problems usually met by alternative algorithms.
\item Feature vector comparison is an easy binary operation.
\item Feature vectors' length depends on $N$. A small $\mbox{DB}_R$ is less expensive in terms of memory. In fact if $\rho$  is small, we won't need a lot of representative vicinities to cover all possibilities. If $\rho$ is large, we will need more representative vicinities, but achieve a better matching accuracy.
\end{itemize}

We now show how to improve this algorithm in two different ways.

\section{Second-Order Vicinities}

The algorithm that we have just described is insensitive to the relative positions of vicinities. {\sl i.e.} two fingerprints containing the same vicinities at {\sl different} locations will be considered equivalent. It is hence natural to try to squeeze more information out of point constellations by considering the relative positions of vicinities.\smallskip

{\sl Second-order vicinities} are defined as vicinities of vicinities. Run the algorithm as before and extract (first-order) vicinities. Replace each vicinity by its barycenter and get a new scatter plot. Group each plot into new vicinities, larger than previous ones. These new vicinities are our {\sl second-order vicinities}. This takes into account the position of a vicinity with respect to the other vicinities and hence eases the discrimination of two constellations containing identical vicinities at different places.\\

The program doing this, available from the authors, computes vicinities as before and then filters them to keep only "significant" vicinities, having more than $\ell_{\mbox{{\scriptsize min}}}$ and less than $\ell_{\mbox{{\scriptsize max}}}$ vicinities. We delete vicinities whose central minutiae belong to other vicinities to keep vicinities pairwise distinct. Once done, we reduce the vicinity representation to their central minutiae. Finally, we run the BD algorithm, except that we now use the vicinity barycenters instead of the center minutiae, and a radius $\rho_2$ larger than the $\rho_1$ used for the first-order vicinities. We can notice that at first sight, the final representation of second-order vicinities is not that different from a larger first-order vicinity. The difference is in the recognition. In fact, we make a double pass recognition, one for first-order vicinities (in second-order vicinities) and one for minutiae (in first-order vicinities). The same algorithm is hence run twice with different parameters. A fingerprint will match if both its first-order and second-order vicinities match the target template. With proper parameter tuning, this algorithm experimentally improves recognition accuracy.\\

Higher order vicinities can be defined as well. But the higher is the order, the larger will the (higher-order) vicinity be. Hence, the information increment brought by higher order vicinities quickly decreases with the order.

\section{Missing Minutiae Analysis}

\begin{figure}[h]
\centering
	\includegraphics[scale=0.8]{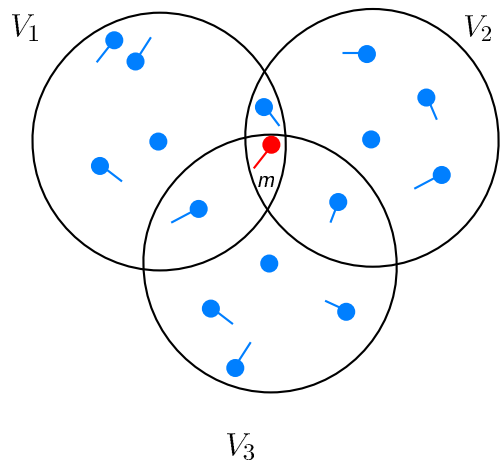}
    \caption{Example of a Missing Minutia}
    \label{Missing minutiae}
\end{figure}

Let's go back to BD's classical (first-order) vicinities. We will now extract information from accidentally missing minutiae. Inferring the missing minutiae and taking them into account reduces the algorithm's False Reject Rate. As shown in Figure \ref{Missing minutiae}, a given minutia may simultaneously belong to ({\sl i.e.} be at the intersection of) several vicinities.\\

If several candidate vicinities present a missing minutia with respect to their template vicinities, we may suspect that a minutia common to these vicinities was omitted accidentally. In other words the {\sl simultaneous disappearance} of $m$ in Figure \ref{Missing minutiae} would be a "smoking gun" indicating an "explainable error". The computational strategy implementing this intuitive observation turns out to be a nontrivial exercise\footnote{The Mathematica code is available from the authors.}.\\

The Hungarian algorithm compares vicinities (hereafter $A$ and $B$) pairwise. To do so, the Hungarian algorithm takes as input a square matrix containing the matching scores $s(a_i, b_j)$ computed between the minutiae of $A$ (lines) and $B$ (columns). If $B$ has less minutiae than $A$, a virtual column is created. All the elements of this virtual column are set to the value $\max_{i,j} s(a_i, b_j)$. After associations are found, the associations corresponding to the virtual column are removed from the list of associations between the minutiae of $A$ and $B$.\\

The score $S(A, B)$ is the sum of the matrix elements corresponding to the associations, plus a penalty for non-associated elements. If the associated minutiae match each other, $S(A, B)$ is essentially equal to the penalty. In that case, we can plausibly assume that one of the vicinities ($A$ or $B$) presents a missing minutia.\\

This scheme can be generalized to more missing minutiae but claims time exponential in the number of missing minutiae. It is hence useful for assessing the occlusion (disappearance) of one or two minutiae only.

\section{Can Nuts \& Bolts Compare?}

This section presents a rather unusual point constellation recognition algorithm. One advantage of this scheme is the fact that it can be easily implemented using physical engines. A physics engine is a computer software that provides an approximate simulation of certain physical systems, such as rigid body dynamics (including collision detection), soft body dynamics, and fluid dynamics, of use in the domains of computer graphics, video games and films. Physical engines are mainly used in video games (typically as middleware), in which case the simulations are in real-time.\smallskip

Let $a_i=(a_{i,x},a_{i,y}) \in \mathbb{R}^2$ and $b_i=(b_{i,x},b_{i,y}) \in \mathbb{R}^2$ denote points in $\mathbb{R}^2$. The reader may consider each such point as the barycenter of a specific vicinity or as the $\{x_i,y_i\}$ coordinates of minutiae to compare.\smallskip

Let $\mathcal{A}=\{a_1,\ldots,a_\ell\}$ and $\mathcal{B}=\{b_1,\ldots,b_\ell\}$ be two sets of points.\smallskip

Our goal is to {\sl measure the geometrical similarity} between the point constellations $\mathcal{A}$ and $\mathcal{B}$.

Define the operator $\Gamma$ by:

\[a'_i=\Gamma(a_i,\Delta{x},\Delta{y},\theta)=
\left(\begin{array}{c}
a'_{i,x}\\
a'_{i,y}\\
\end{array}
\right)
=
\left( \begin{array}{cc}
\cos \theta & - \sin \theta \\
\sin \theta & \cos \theta  \end{array} \right)
\left(\begin{array}{c}
a_{i,x}\\
a_{i,y}\\
\end{array}
\right)+
\left(\begin{array}{c}
\Delta{x}\\
\Delta{y}\\
\end{array}
\right)\]

In other words, $\Gamma$ translates a point by $\Delta{x},\Delta{y}$ and rotates it by the angle $\theta$.\smallskip

We define the {\sl similarity} between the constellations $\mathcal{A}$ and $\mathcal{B}$ by:

$$\mbox{sim}(\mathcal{A},\mathcal{B})= \min_{\Delta{x},\Delta{y},\theta} \left(\sum_{i=1}^\ell \mbox{dist}(\Gamma(a_i,\Delta{x},\Delta{y},\theta),b_i)\right)\in \mathbb{R}$$

Where $\mbox{dist}((x,y),(x',y'))=\sqrt{(x-x')^2+(y-y')^2}$.\smallskip

Informally, $\mbox{sim}(\mathcal{A},\mathcal{B})$ measures the total distance remaining between the points of $\mathcal{A}$ and $\mathcal{B}$ after the best possible translation and rotation attempting to match $\mathcal{A}$ and $\mathcal{B}$. It is easy to see that $0\leq\mbox{sim}(\mathcal{A},\mathcal{B})<\infty$ where $\mbox{sim}(\mathcal{A},\mathcal{B})=0 \Leftrightarrow \mathcal{A}=\mathcal{B}$ (perfect match).\smallskip

In applications requiring a $[0,1]$ score, one can use $c^{-\mbox{{\scriptsize sim}}}$ for some constant $c>1$ ({\sl e.g.} the base of natural logarithms $e$).\smallskip

The analytical determination of $\mbox{sim}(\mathcal{A},\mathcal{B})$ turns out to be a complex task. We hence replace $\mbox{sim}$ by a function $\mbox{sim}_{\phi}$ which effect approximates $\mbox{sim}$. As will be explained in the next section, the $\phi$ in $\mbox{sim}_{\phi}$ denotes the fact that this new similarity measurement function is inspired by {\sl physics}.

\subsection{Elastic Potential Energy Matching}

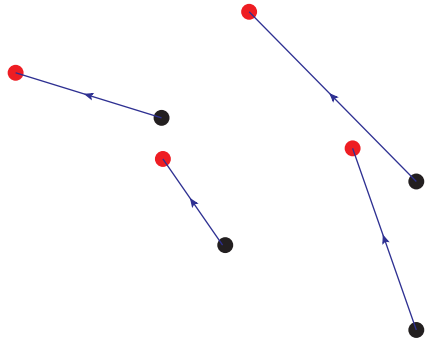
\begin{figure}
\centering
\scalebox{.5}{
  \begin{picture}(304,267)(0,27)%(316,130)(0,39)
    \SetWidth{4.0}
    \SetColor{Red}
    \SetWidth{1.0}
    \Vertex(113,156){6}
    \Vertex(2,221){6}
    \Vertex(256,164){6}
    \Vertex(178,267){6}
    \SetWidth{4.0}
    \SetColor{Black}
    \SetWidth{1.0}
    \Vertex(112,187){6}
    \Vertex(160,91){6}
    \Vertex(304,139){6}
    \Vertex(304,27){6}
        \SetColor{Blue}
    \Line[arrow,arrowpos=0.5,arrowlength=5,arrowwidth=2,arrowinset=0.2](112,187)(2,221)
        \Line[arrow,arrowpos=0.5,arrowlength=5,arrowwidth=2,arrowinset=0.2](304,139)(178,267)
            \Line[arrow,arrowpos=0.5,arrowlength=5,arrowwidth=2,arrowinset=0.2](304,27)(256,164)
                \Line[arrow,arrowpos=0.5,arrowlength=5,arrowwidth=2,arrowinset=0.2](158,91)(113,156)
  \end{picture}}
\caption{Two Point Constellations (Black \& Red) to Match} \label{fig:M1}
\end{figure}

\begin{figure}
\centering
\scalebox{.5}{
  \begin{picture}(304,267)(0,27)%(316,130)(0,39)%
    \SetWidth{1.0}
    \SetColor{Blue}
    \Gluon(116,194)(112,189){7.5}{5} %OK
    \Gluon(180,82)(160,93){7.5}{5} %OK
    \Gluon(292,146)(304,141){7.5}{5}
    \Gluon(308,18)(304,29){7.5}{5}
    \SetWidth{4.0}
    \SetColor{Red}
    \Line(116,194)(292,146)
    \Line(180,82)(308,18)
    \Line(292,146)(308,18)
    \Line(116,194)(180,82)
    \SetWidth{1.0}
    \Vertex(180,82){6}
    \Vertex(116,194){6}
    \Vertex(308,18){6}
    \Vertex(292,146){6}
    \SetWidth{4.0}
    \SetColor{Black}
    \Line(112,189)(160,93)
    \Line(160,93)(304,29)
    \Line(112,189)(304,141)
    \Line(304,141)(304,29)
    \SetWidth{1.0}
    \Vertex(112,189){6}
    \Vertex(160,93){6}
    \Vertex(304,141){6}
    \Vertex(304,29){6}
  \end{picture}
}
\caption{Rigidified Constellations Attached with Springs Stabilized After Release} \label{fig:M3}
\end{figure}
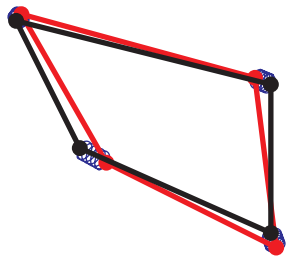

The two point constellations $\mathcal{A}$ and $\mathcal{B}$ are modeled as fully rigid physical solids (illustrated by the red and black objects in Figure \ref{fig:M1}). We link each couple of related points (i.e. $a_i$ and $b_i$) with physical springs (Figure \ref{fig:M2}), release the objects and simulate the resulting evolution of the mechanical system using linear integral interpolation to get the result shown in Figure \ref{fig:M3}.\smallskip

Recall that elastic potential energy is the potential energy stored following the deformation of a spring. This energy is the work necessary to stretch the spring by $x$ length units. According to Hooke's law, the force $F=-kx$ required to stretch a spring is proportional to $x$. Since the change in spring's potential energy between two positions is the work required to move the object from point 0 to point $x$, the spring's potential energy at distance $x$ is $\frac{kx^2}{2}$. We assume that the springs used in our model start extending from a length of zero.\smallskip

With well-chosen parameters, the system converges quickly to a minimal-energy position (measured by the potential energy stored by the springs). The higher this stable energy is, the further apart the sets are.\smallskip

\begin{figure}[h]
\centering
	\includegraphics[scale=1]{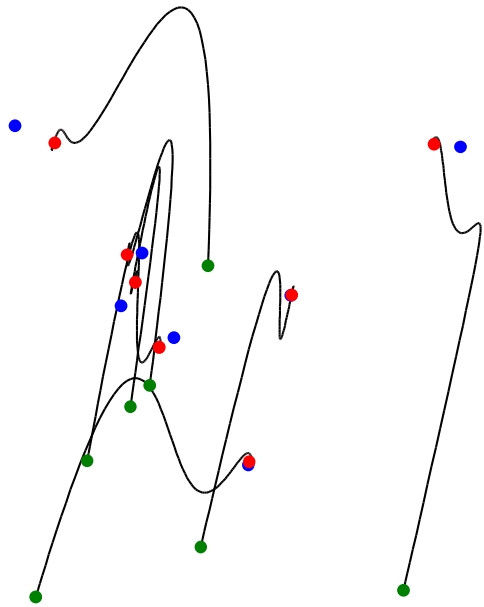}
    \caption{Convergence Plot}
    \label{convergence}
\end{figure}

Figure~\ref{convergence} shows the convergence between the red (candidate vicinities $\mathcal{A}$) and the blue plots (template vicinities $\mathcal{B}$). The green plots represent the initial position of $\mathcal{A}$ before the release of the system. The blue plots are assigned an infinite mass and hence don't move. We can see in this figure that after the release of the system, the red plots are really close (and at times even superimposed on) to blue plots. The system's potential energy or the sum of the distances between red and blue plots after system convergence provide two possible similarity scores.\\

Let $\mathcal{A}'$ be the solid obtained by placing a fixed unitary mass $m_{a,i}$ at each of the $\ell$ coordinates given by $\mathcal{A}$. All the unitary masses of $\mathcal{A}'$ are linked together by weightless rigid bars. Exactly the same construction is applied to $\mathcal{B}$ to yield a mechanical structure $\mathcal{B}'$ composed of infinite masses $m_{b,i}$. Fixing $m_{b,i}=\infty$ makes $\mathcal{B}'$ immovable.\smallskip

We then add for all $i$ a spring linking $m_{a,i} \leftrightsquigarrow m_{b,i}$ and release the system. Following this release, $\mathcal{A}'$ and $\mathcal{B}'$ will move to a position minimizing the sum of spring energies, while preserving the spatial offsets between the mass points composing $\mathcal{A}'$ and $\mathcal{B}'$. \smallskip

We can manipulate two parameters : the strength coefficient of the spring, and the friction coefficient. The friction force we apply is $F=-k_v V$ where $V$ is the speed. If we choose a friction coefficient which is too low, the springs become too strong and the system keeps oscillating for too long. If we choose a too low force coefficient, the figures will never match in reasonable time. By using well-chosen parameters, the system stabilizes in less than a second. After defining a proper stop condition ({\sl cf. infra}), we get Algorithm 1.\smallskip

\begin{algorithm}
\caption{The Mechanical Comparator}
\begin{algorithmic}[1]
	\STATE ``Assemble'' rigid objects $\mathcal{A}'$ and $\mathcal{B}'$ from the point sets $\mathcal{A}$ and $\mathcal{B}$ assigning an infinite mass to $\mathcal{B}'$.
	\STATE Compute the moment of inertia of $\mathcal{A}'$
    	\WHILE {stop condition is not met}
            \STATE Compute springs forces and moments
            \STATE Compute the next position of $\mathcal{A}'$ using integral linear interpolation
        \ENDWHILE
\end{algorithmic}
\end{algorithm}

\begin{figure}
\centering
\scalebox{.5}{
  \begin{picture}(304,267)(0,27)%(316,130)(0,39)
    \SetWidth{1.0}
    \SetColor{Blue}
    \Gluon(112,187)(2,221){7.5}{5}
    \Gluon(304,139)(178,267){7.5}{5} %OK
    \Gluon(304,27)(256,164){7.5}{5} %OK
    \Gluon(158,91)(113,156){7.5}{5}
    \SetWidth{4.0}
    \SetColor{Red}
    \Line(2,221)(178,267)
    \Line(113,156)(256,164)
    \Line(178,267)(256,164)
    \Line(2,221)(113,156)
    \SetWidth{1.0}
    \Vertex(113,156){6}
    \Vertex(2,221){6}
    \Vertex(256,164){6}
    \Vertex(178,267){6}
    \SetWidth{4.0}
    \SetColor{Black}
    \Line(112,187)(160,91)
    \Line(160,91)(304,27)
    \Line(112,187)(304,139)
    \Line(304,139)(304,27)
    \SetWidth{1.0}
    \Vertex(112,187){6}
    \Vertex(160,91){6}
    \Vertex(304,139){6}
    \Vertex(304,27){6}
  \end{picture}
}
\caption{Rigidified Constellations Attached with Springs Before Release} \label{fig:M2}
\end{figure}
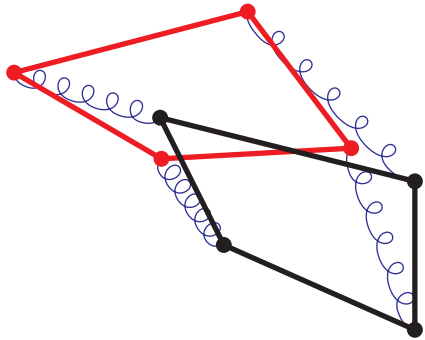

For now, the only stop condition that we use is the stabilization of the system, but as further improvements we may consider having other stop conditions.\\

\begin{figure}[!h]
\centering
	\includegraphics[scale=0.5]{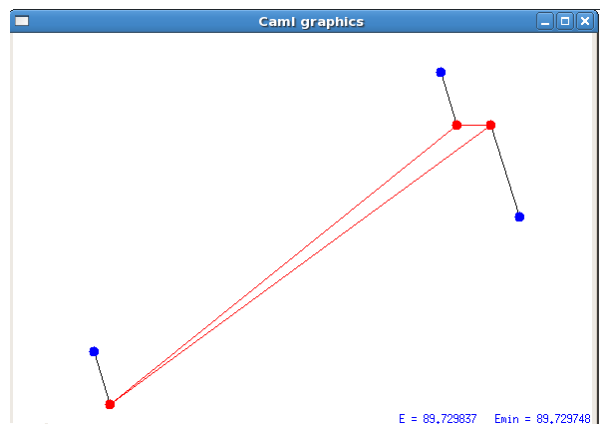}
    \caption{The Mechanical Matching Software}
    \label{spring}
\end{figure}

The software (Figure \ref{spring}), developed in Ocaml, is available from the authors. The blue plots represent the (infinite mass) template minutiae and the red plots represent the (mobile) candidate minutiae. The red lines represent the rigid structure formed by the candidate minutiae. Black lines represent springs. $E$ shows the system's energy at time $t$ and $E_{\mbox{{\scriptsize min}}}$ is the minimum potential energy released by the system.

\subsection{Further Research}

This work opens a number of interesting research perspectives:

\begin{itemize}
   \item Is it possible to leverage physical attraction-repulsion phenomena to build concrete {\sl analog} point constellation comparison hardware? {\sl i.e.} cheap and disposable analog image recognition coprocessors? Because opposite electrical particles essentially act as springs this might not be impossible. However, it remains to somehow neutralize the "crosstalk attraction" between unrelated points ({\sl i.e.} the attraction of $a_i$ by $b_j$ for $i\neq j$). Note that such {\sl physically built} systems can also serve as real-life alert systems: by equipping a bridge with sensors and consolidating the global system energy, an alert can be launched if the mechanical strain on the bridge becomes dangerously high.
   \item Add to the simulation system perturbations to avoid meta-stable states. Experimentally, such meta-stable states appeared when extreme physical coefficient values were chosen.
   \item Match point-sets more accurately (by making the moving system partially flexible. This will capture the intuition that a finger or a face are soft masses whose forms could have slightly been stretched if we wanted to. Hence, by allowing the mobile object some flexibility we reproduce a slight "grimace" that the user could have naturally done if we would have asked him to. Flexibility can be naturally implemented by embedding in each solid bar an inner spring or by replacing edges by mechanical articulations with two degrees of freedom.
   \item Matching non-tagged points with multiple springs. Here the idea consists in finding the proper assignment between the $a_i$ and the $b_j$ using multiple springs before comparing the tagged point constellations. We note that if $\mathcal{A}$ and $\mathcal{B}$ only differ by an unknown rotation (of a known center), a forced rotation of $\mathcal{A}$ during which we monitor the system's potential energy will reveal the unknown angle by a sudden global system energy drop. While this amounts to exhaustively searching the unknown angle, it is probable that coarse-grain interpolation could be used to avoid testing useless configurations. For instance one possibility would be to sample angles with a $10^{\circ}$ increment, select those having the smallest energy and refine the search. We did not try this so far and it is pretty probable that artificially crafted (or maybe naturally encountered objects) would fail to be detected. Finding and/or approximating the optimal $\Delta{x},\Delta{y}$ using this method is another open problem.
   \item Optimize friction, force and mass coefficients as a function of correlation between $\mbox{sim}_{\phi}$ and $\mbox{sim}$ using simulated annealing. Simulated annealing is a generic probabilistic metaheuristic for the global optimization problem of locating a good approximation to the global optimum of a given function in a large search space. It is often used when the search space is discrete ({\sl e.g.}, all tours that visit a given set of cities). For certain problems, simulated annealing may be more efficient than exhaustive enumeration — provided that the goal is merely to find an acceptably good solution in a fixed amount of time, rather than the best possible solution. This seems to be case here. One possibility could be to generate a library of objects $\mathcal{A}_i$, slightly distort the $\mathcal{A}_i$s to obtain $\mathcal{B}_i$s, randomly rotate and translate the $\mathcal{B}_i$s and compute the Pearson correlation coefficient $\rho[\mbox{sim}_{\phi},\mbox{sim}]$ between the data sets $\{\mbox{sim}_{\phi}(\mathcal{A}_i,\mathcal{B}_i)\}$ and $\{\mbox{sim}(\mathcal{A}_i,\mathcal{B}_i)\}$. Using simulated annealing determine the friction, force and mass coefficients that maximize $\rho[\mbox{sim}_{\phi},\mbox{sim}]$.
   \item The viscosity of a fluid is a measure of its resistance to gradual deformation by shear stress or tensile stress. For liquids, it corresponds to the informal notion of "thickness". For example, honey has a higher viscosity than water. Viscosity is due to the friction between neighboring particles in a fluid that are moving at different velocities. When the fluid is forced through a tube, the fluid generally moves faster near the axis and very slowly near the walls; therefore, some stress (such as a pressure difference between the two ends of the tube) is needed to overcome the friction between layers and keep the fluid moving. For the same velocity pattern, the stress required is proportional to the fluid's viscosity. A liquid's viscosity depends on the size and shape of its particles and the attractions between the particles. Replacing mechanical friction by viscosity ({\sl i.e.} simulating the behavior of the system immersed in water or honey) can be explored.
  \item Finally, apply this approach to image face recognition (the characteristic points of the face are known and easy to detect and are thus easy to match).
\end{itemize}

\bibliographystyle{plain}
\bibliography{biblio}
\end{document}